\documentclass[sigconf]{acmart}
\AtBeginDocument{%
  }
\copyrightyear{2025}
\acmYear{2025}
\setcopyright{acmlicensed}\acmConference[MM '25]{Proceedings of the 33rd ACM International Conference on Multimedia}{October 27--31, 2025}{Dublin, Ireland}
\acmBooktitle{Proceedings of the 33rd ACM International Conference on Multimedia (MM '25), October 27--31, 2025, Dublin, Ireland}
\acmDOI{10.1145/3746027.3758290}
\acmISBN{979-8-4007-2035-2/2025/10}
\begin{document}

%\title{\textcolor{red}{[fangyu: please draft some titles, we can select one.]}}
% \title{FEdit: Constructing a Large Scale Fashion Image Editing Dataset with Automated Pipeline and Fine-Grained Evaluation}
% \title{FEdit: A large scale dataset and Semantic Evaluation Framework for Instruction-Based Garment Image Editing}
\title[EditGarment: An Instruction-Based Garment Editing Dataset]{EditGarment: An Instruction-Based Garment Editing Dataset Constructed with Automated MLLM Synthesis and Semantic-Aware Evaluation}

\author{Deqiang Yin}
\email{Deqiang.Yin24@student.xjtlu.edu.cn}
\authornote{Equal Contribution.}
\affiliation{%
  \institution{Xi'an Jiaotong-Liverpool University}
  \city{Suzhou}
  \state{Jiangsu}
  \country{China}
}
\author{Junyi Guo}
\authornotemark[1]
\email{Junyi.Guo23@student.xjtlu.edu.cn}
\affiliation{%
  \institution{Xi'an Jiaotong-Liverpool University}
  \city{Suzhou}
  \state{Jiangsu}
  \country{China}
}
\author{Huanda Lu}
\email{huandalu@nbt.edu.cn}
\affiliation{%
  \institution{NingboTech University}
  \city{Ningbo}
  \state{Zhejiang}
  \country{China}
}
\author{Fangyu Wu}
\authornote{Corresponding Author.}
\email{Fangyu.Wu02@xjtlu.edu.cn}
\affiliation{%
  \institution{Xi'an Jiaotong-Liverpool University}
  \city{Suzhou}
  \state{Jiangsu}
  \country{China}
}
\author{Dongming Lu}
\email{ldm@zju.edu.cn}
\affiliation{%
  \institution{Zhejiang University}
  \city{Hangzhou}
  \state{Zhejiang}
  \country{China}
}


\renewcommand{\shortauthors}{Deqiang Yin, Junyi Guo, Huanda Lu, Fangyu Wu, \& Dongming Lu}

\begin{abstract}
% Instruction-based garment editing enables precise image modifications via natural language, with broad applications in fashion design and customization. Unlike general editing tasks, it requires understanding garment-specific semantics and attribute dependencies. However, progress is limited by the scarcity of high-quality instruction–image pairs, as manual annotation is costly and hard to scale. While MLLMs have shown promise in automated data synthesis, their application to garment editing is constrained by imprecise instruction modeling and a lack of fashion-specific supervisory signals. 
% To address these challenges, we present an automated pipeline for constructing a garment editing dataset. We first define six editing instruction categories aligned with real-world fashion workflows to guide the generation of balanced and diverse instruction–image triplets.
% Second, we introduce Fashion Edit Score, a semantic-aware evaluation metric that captures semantic dependencies between garment attributes and provides reliable supervision during construction. Based on this pipeline, we present EditGarment, a dataset of 51,371 high-quality triplets, the first instruction-based dataset tailored to standalone garment editing. Experiments show that EditGarment provides a baseline result for the garment editing task. We provide sample data, and the complete dataset will be available at https://anonymous.4open.science/r/EditGarment/.

Instruction-based garment editing enables precise image modifications via natural language, with broad applications in fashion design and customization. Unlike general editing tasks, it requires understanding garment-specific semantics and attribute dependencies. However, progress is limited by the scarcity of high-quality instruction–image pairs, as manual annotation is costly and hard to scale. While MLLMs have shown promise in automated data synthesis, their application to garment editing is constrained by imprecise instruction modeling and a lack of fashion-specific supervisory signals.  To address these challenges, we present an automated pipeline for constructing a garment editing dataset. We first define six editing instruction categories aligned with real-world fashion workflows to guide the generation of balanced and diverse instruction–image triplets. Second, we introduce Fashion Edit Score, a semantic-aware evaluation metric that captures semantic dependencies between garment attributes and provides reliable supervision during construction. Using this pipeline, we construct a total of 52,257 candidate triplets and retain 20,596 high-quality triplets to build EditGarment, the first instruction-based dataset tailored to standalone garment editing. The project page is https://yindq99.github.io/EditGarment-project/.

\end{abstract}

%    <concept>
%        <concept_id>10010147.10010371.10010382.10010383</concept_id>
%        <concept_desc>Computing methodologies~Image processing</concept_desc>
%        <concept_significance>300</concept_significance>
%        </concept>
%  </ccs2012>
% \ccsdesc[300]{Computing methodologies~Image processing}

\begin{CCSXML}
<ccs2012>
   <concept>
       <concept_id>10010147.10010178.10010224.10010240</concept_id>
       <concept_desc>Computing methodologies~Computer vision representations</concept_desc>
       <concept_significance>500</concept_significance>
       </concept>
 </ccs2012>
\end{CCSXML}

\ccsdesc[500]{Computing methodologies~Computer vision representations}


\keywords{Instruction-based garment editing, Automatic dataset construction}

\begin{teaserfigure}
  \centering
  \includegraphics[width=0.8\textwidth]{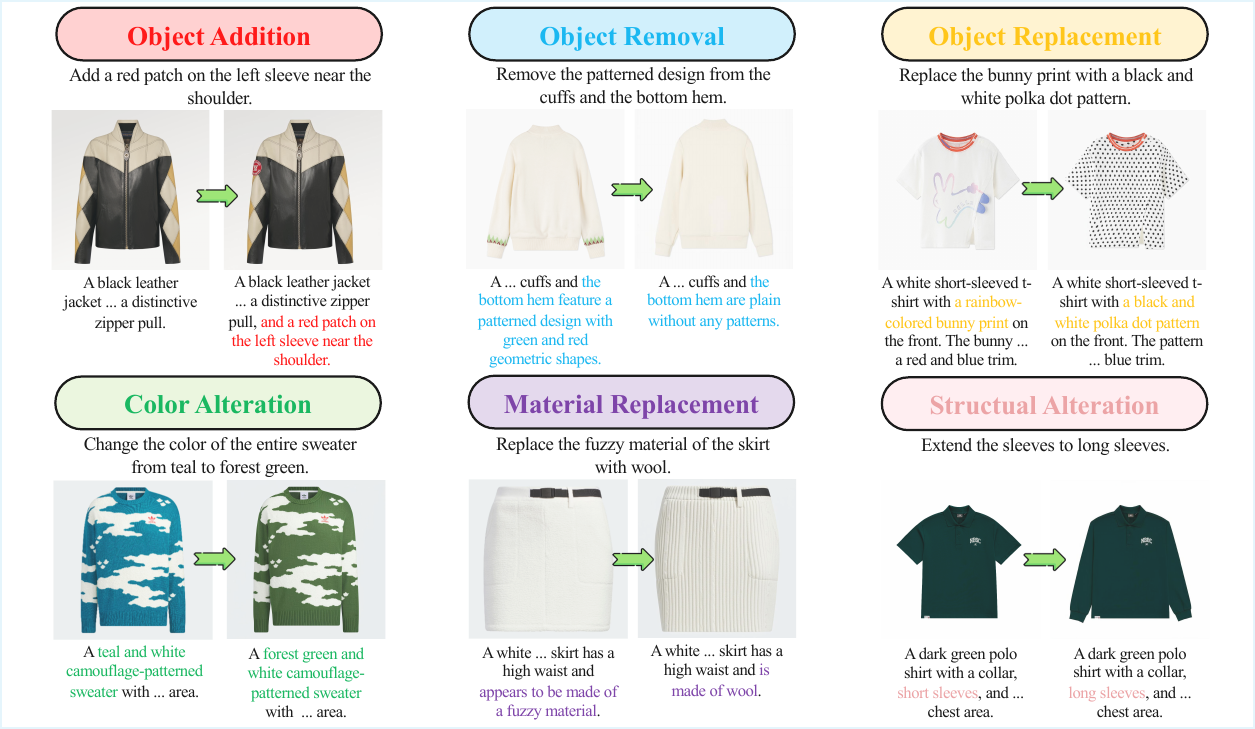}
  \caption{Example images and edit instructions from EditGarment; our dataset defines six editing types corresponding to common professional fashion design needs.}
  %\Description{Enjoying the baseball game from the third-base
  %seats. Ichiro Suzuki preparing to bat.}
  \label{fig:teaser}
\end{teaserfigure}


\maketitle

\section{Introduction}

Instruction-based image editing (IIE) is an emerging multimodal task that aims to modify images based on natural language instructions, enabling fine-grained, controllable visual editing \cite{brooks2023instructpix2pix,huberman-spiegelglasEditFriendlyDDPM2024,mokadyNulltextInversionEditing2023,huangSmartEditExploringComplex2024,liuUnderstandingCrossSelfAttention2024,tewelAdditTrainingFreeObject2024,brackLEDITSLimitlessImage2024,bodurIEditLocalisedTextguided2024,muEditARUnifiedConditional2025,sheyninEmuEditPrecise2024}. Its potential has attracted growing interest due to a wide range of applications in domains such as e-commerce, advertising, and digital designing \cite{shuaiSurveyMultimodalGuidedImage2024,huang2025diffusion}.

As a domain-specific subtask, garment editing involves modifying clothing images according to user-provided instructions, demanding fine-grained control over garment components and attributes. By enabling repeated and controllable visual modifications, it has the potential to significantly accelerate iterative fashion design and reduce the cost associated with traditional prototyping cycles.

However, instruction-based garment editing remains relatively underexplored in the literature yet. Most existing works in the fashion domain focus on garment generation \cite{guo2025higarmentcrossmodalharmonybased,zhangDiffClothDiffusionBased2023,zhangGarmentAlignerTexttoGarmentGeneration2024,zhangARMANIPartlevelGarmentText2022} or virtual try-on \cite{baldratiMultimodalGarmentDesigner2023,baldratiMultimodalConditionedLatentDiffusion2024,wangTexFitTextDrivenFashion2024,morelliLaDIVTONLatentDiffusion2023,morelliDressCodeHighResolution2022, VITON-HD_Choi_2021_CVPR}, which do not support instruction-driven modifications to standalone garment images. In addition, the absence of high-quality datasets that pair detailed editing instructions with corresponding image pairs presents a substantial bottleneck. Manual annotation of such datasets is both labor-intensive and costly, limiting scalability and impeding the development of robust and generalizable garment editing models. To fill this gap, this work aims to automatically construct a large-scale dataset tailored for instruction-based garment editing. 

The challenge of automatic dataset construction in image editing task was first tackled by InstructPix2Pix \cite{brooks2023instructpix2pix}, which used GPT-3 \cite{brown2020language} and Stable Diffusion \cite{rombach2022high} to generate paired instructions and images. Subsequent studies have attempted to improve image resolution and edit–instruction alignment by incorporating human feedback \cite{zhangHIVEHarnessingHuman2024} or segmentation masks \cite{zhangMAGICBRUSHManuallyAnnotated}. HQ-EDIT \cite{huiHQEditHighQualityDataset2024} further devised a scalable data collection pipeline leveraging advanced foundation models, namely GPT-4V and
DALL-E 3. Despite recent advances in automated dataset construction using MLLMs, existing methods often fall short when applied to instruction-based garment editing. Specifically, many approaches fail to generate diverse and semantically accurate edits aligned with fine-grained instructions; moreover, they lack reliable evaluation signals tailored to the fashion domain, making it difficult to ensure the quality of generated data.

To address these limitations, effective instruction-based garment editing datasets must overcome two key challenges:
\textbf{(1) Long-tail attribute coverage:} Existing MLLMs are biased toward common garment modifications due to their pretraining on general-domain data, and struggle with rare but important edits such as fabric replacement or subtle style changes.
\textbf{(2) Semantics-aware supervision:} Traditional metrics like SSIM \cite{SSIM} or CLIP similarity \cite{CLIPradfordLearningTransferableVisual2021} are insufficient for verifying whether the instructed changes are faithfully reflected, especially at the attribute level. A reliable, interpretable evaluation signal is required to ensure both edit fidelity and visual consistency. When considering both challenges, current pipelines still show a significant gap in supporting high-quality dataset construction for garment editing tasks.

To build a dataset with balanced distribution across editing categories and reliable semantics-aware supervision, we propose a fully automated MLLM-based pipeline. To mitigate the long-tail distributions produced by MLLM, we design an auto-synthesis framework. This framework consists of six prompt templates with corresponding triplets—original description, edit prompt, and edited description that contains six common editing categories. These categories reflect real-world production workflows and cover the most common editing operations used by fashion professionals. These templates ensure balanced coverage of practical editing scenarios, guiding the MLLM to generate a diverse and uniformly distributed dataset of garment editing examples. %By training on this balanced data, the fine-tuned editing model acquires comprehensive domain-specific priors and fine-grained control over garment attributes. 
To provide a semantic-aware supervision signal for the automated pipeline, we propose Fashion Edit Score (FEditScore). FEditScore builds a semantic dependency graph composed of structured yes/no questions, further computing a score to evaluate the quality of edit operations. Building on the existing garment generation dataset \cite{guo2025higarmentcrossmodalharmonybased}, we augmented it with our pipeline to construct EditGarment. EditGarment contains 20,596 high quality editing triplets, and some examples are shown in Fig. \ref{fig:teaser}.

Our contributions are summarized as follows:
\begin{itemize}
\item We build EditGarment, the first large-scale multi-modal dataset for the garment editing task. EditGarment contains 20,596 edits of original garment images, edit instructions, and edited images across diverse categories.

\item We introduce a fully automated MLLM-based pipeline that leverages six edit categories reflecting real-world fashion design needs, enabling balanced and diverse triplet generation on garment images.

\item We propose Fashion Edit Score (FEditScore), a novel evaluation metric that uses a semantic dependency graph to capture and evaluate fine-grained edit operations in garment image editing.
\end{itemize} 

\section{Related Work}

\subsection{Image Editing Datasets}
Given the inherent complexity of pairing visual content with detailed editing instructions, constructing high-quality datasets for image editing has been a longstanding challenge. Magicbrush \cite{zhangMAGICBRUSHManuallyAnnotated} manually annotates 10k edits utilizing DALL-E 2 \cite{dalle-mishkin2022risks}.
Recently, there have been endeavors to synthesize large-scale image editing datasets. For example, InstructPix2Pix \cite{brooks2023instructpix2pix}, HIVE \cite{zhangHIVEHarnessingHuman2024}, FaithfulEdit \cite{chakrabartyLearningFollowObjectCentric2023a}, and HQ-Edit \cite{huiHQEditHighQualityDataset2024} synthesize instruction–image pairs using the latest large models and generic prompts. However, these methods focus on general editing tasks.

Within image editing research in the fashion domain, several large-scale datasets have been developed for virtual try-on tasks. Such as DressCode \cite{morelliDressCodeHighResolution2022}, which provides 53,795 image–text pairs for garment-to-human fitting, and VITON-HD \cite{VITON-HD_Choi_2021_CVPR}, which includes 13,679 high-resolution aligned pairs. These datasets are primarily designed for pose-guided garment transfer and do not support editing garments independently of the person.

Recently, several datasets for garment generation have been proposed, such as CM-Fashion \cite{zhangARMANIPartlevelGarmentText2022}, which contains 500,000 garment images with corresponding textual descriptions, and MMDGarment \cite{guo2025higarmentcrossmodalharmonybased}, which provides 20,151 text–image pairs with attribute-level annotations such as fabric. However, these datasets are primarily designed for garment synthesis and lack instruction-based editing supervision. In this paper, we present an automated pipeline for constructing the first large-scale dataset specifically designed for instruction-based garment editing.

\subsection{Evaluation Mechanisms for Generated Image}
Recent studies \cite{brooks2023instructpix2pix,huiHQEditHighQualityDataset2024} have leveraged foundation multi-modal models such as Stable Diffusion (SD1.5) \cite{rombach2022high}, DALL-E 3 \cite{dalle-mishkin2022risks}, and GPT series \cite{openaiGPT4TechnicalReport2024} to synthesize image editing datasets, making robust evaluation mechanisms essential for ensuring the quality of generated data in automated annotation pipelines. Traditional evaluation methods for instruction-based image editing (IIE) rely on general visual evaluation metrics such as SSIM \cite{SSIM}, LPIPS \cite{LPIPS-zhang2018unreasonable}, and CLIP similarity \cite{CLIPradfordLearningTransferableVisual2021}, which lack the semantic depth to assess fine-grained garment edits and often fail to reflect whether the visual changes follow the edit instructions.

In text-to-image generation \cite{dhariwalDiffusionModelsBeat,peeblesScalableDiffusionModels2023,rombach2022high}, Visual Question Answering (VQA) has been explored as an auxiliary tool to improve semantic evaluation \cite{cho2023davidsonian,huTIFAAccurateInterpretable2023}.
Notably, FaithfulEdit \cite{chakrabartyLearningFollowObjectCentric2023a} introduced a VQA-based filter to identify unfaithful edits in synthetic datasets. However, its reliability is limited for complex instructions and does not generalize well to garment-specific editing tasks. To improve VQA evaluation, DSG Score \cite{cho2023davidsonian} introduced an empirically grounded framework based on Davidsonian Scene Graphs, effectively addressing reliability issues.
In this work, we discover a novel evaluation mechanism for IIE in the fashion domain by constructing a VQA-based semantic dependency graph. %exploring the semantical interpretability of edit instructions.}

\begin{figure*}
    \centering
    \includegraphics[width=0.75\linewidth]{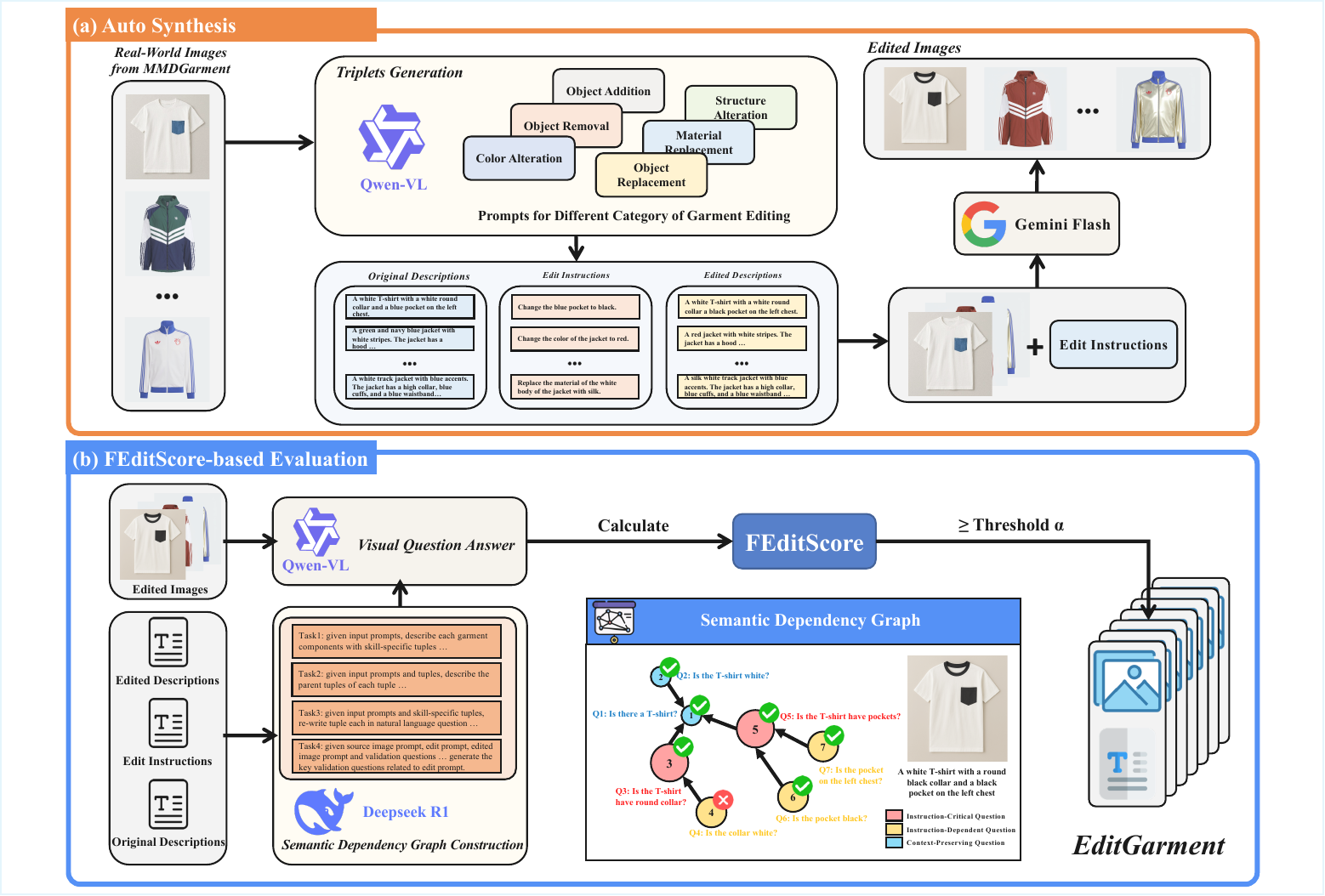}
    \caption{Our dataset construction pipeline contains part (a): first, we apply Qwen-VL \cite{baiQwenVLVersatileVisionLanguage2023} to generate editing triplets regarding our predefined editing categories; and part (b): second, we design a dependency graph structure to quantitatively evaluate the quality of generated data. The semantic dependency graph in the second part illustrates a visualization of our proposed graph structure. Data scored above the predefined threshold will be collected into our EditGarment dataset. }
    \label{fig:2}
\end{figure*}

\section{EditGarment Dataset}
To tackle the garment editing task, we introduce EditGarment, a multi-modal dataset comprising 20,596 triplets of garment images, editing instructions, and the corresponding edited images. As illustrated in Fig. \ref{fig:2}, the EditGarment construction pipeline consists of two stages: Auto Synthesis (Section 3.1) and Fashion Edit Score based Evaluation (Section 3.2). We present the Dataset Analysis in Section 3.3.
To the best of our knowledge, EditGarment is the first publicly available image editing dataset in the fashion domain. 

\subsection{Auto Synthesis}

To ensure the authenticity and diversity of the dataset, we take the real-world garment images sourced from MMDGarment \cite{guo2025higarmentcrossmodalharmonybased} as our original images, which collect a wide spectrum of clothing categories, including tops, bottoms, outerwear, etc. These images serve as the foundation for subsequent transformations, providing visual realism and structural consistency

% These real-world garment images serve as foundational samples, providing the visual and structural grounding for all subsequent semantic and synthetic transformations. 

First, we categorize garment edits into six core types based on the common fashion design practices as follows: %to further strengthen the dataset's expressive range, we construct a taxonomy of six core edit types, each representing a distinct class of fashion modifications:
\begin{itemize}
\item \textbf{Object Removal}: Eliminate specified garment components (e.g., pockets, hoods).
\item \textbf{Object Replacement}: Substitute garment elements with alternative designs.
\item \textbf{Object Addition}: Add new decorative/practical elements.
\item \textbf{Material Replacement}: Alter textile composition (e.g., cotton → silk).
\item \textbf{Color Alteration}: Modify color while preserving texture.
\item \textbf{Structural Alteration}: Reshape the silhouette (e.g., flare → straight cut).
\end{itemize}
This classification framework ensures comprehensive and systematic coverage of plausible garment edits. Each edit type corresponds to a common real-world design operation. 

Second, we employ Qwen-VL \cite{baiQwenVLVersatileVisionLanguage2023}, a state-of-the-art vision-language architecture wth superior cross-modal understanding, to generate triplets for each garment image, following a similar approach to InstructPix2Pix \cite{brooks2023instructpix2pix}. These triplets establish a clear link between the visual input, the editing instruction, and the target output, enabling precise supervision of each edit operation. Each triplet consists of three components: 

\begin{itemize}
\item \textbf{Original Description}: A comprehensive, attribute-rich description of the original garment, capturing its key visual and functional elements (e.g., A gray turtleneck sweater with a ribbed texture). 
\item \textbf{Edit Instruction}: A command that specifies the exact change (e.g., Change the color of the sweater to blue).
\item \textbf{Edited Description}: Specification of target edited image (e.g., A blue turtleneck sweater with a ribbed texture). 
\end{itemize}

This triplet definition not only enables direct mapping between vision and language before and after an edit, but also facilitates the verification of edited images generation in the later stage. After obtaining the editing instructions and descriptions, the next step is the generation of edited images that match the textual data. Finally, we employ Gemini-2.0-Flash \cite{team2023gemini} to generate edited images, which is a powerful image generation model supporting image editing tasks based on textual and visual inputs.

\subsection{Fashion Edit Score based Evaluation}

Automated pipelines with LLM often yield samples with inconsistent or unintended modifications. Unfortunately, an effective detection mechanism is lacking, as existing evaluation metrics demonstrate limited capacity for evaluating fine-grained instruction-image consistency for garment editing.

To address these shortcomings, we introduce fashion edit score (FEditScore) based evaluation mechanism for garment image editing. Inspired by the Davidsonian Scene Graph Score (DSG Score) \cite{cho2023davidsonian}, our evaluation constructs a hierarchical graph of atomic questions to capture the semantic relationships between the \textbf{Edited Image} and the \textbf{Edit Instruction} from Section 3.1. %FEditScore measures the semantic dependencies between the \textbf{Edited Image} and the \textbf{Edit Instruction} from Section 3.1 through a hierarchical semantic dependency graph composed of atomic questions. 
It takes a scoring mechanism that prioritizes instruction-relevant edits in fashion-specific contexts, ensuring accurate evaluation of both edited and preserved regions. Specifically, the evaluation comprises three stages: dependency graph construction, answer generation, and final score calculation. 

\paragraph{Dependency Graph Construction}
In this stage, we build a hierarchical semantic dependency graph through four sequential steps by applying DeepSeek R1\cite{guo2025deepseek} due to its powerful visual understanding potential. First, we perform component-level parsing, where the \textbf{Edited Descriptions} are analyzed to extract all relevant garment components (e.g., sleeves, collars, and pockets) and their associated attributes (e.g., color, length, and fabric). %These components, such as sleeves, collars, and pockets, form the foundation units for the following question generation stage. 
Second, we establish semantic-structural dependencies among these components and attributes according to \textbf{Edited Descriptions}, constructing a fashion-specific graph where nodes represent visual elements and edges encode compositional or dependency relationships. Third, we generate a set of atomic binary questions from this graph dependency. Fourth, we classify and annotate each atomic question into one of three categories: Instruction-Critical Questions (ICQ), Instruction-Dependent Questions (IDQ), or Context-Preserving Questions (CPQ). Based on the semantic relation to the \textbf{Edit Instruction}, we assign each the appropriate weight for the final scoring stage. 

Specifically, we define each of these questions as follows:

\noindent \textbf{Instruction-Critical Questions (ICQ):} These questions directly correspond to the critical semantic elements described in the \textbf{Edit Instructions}. They serve as primary indicators of whether the core editing objective has been achieved successfully. ICQs are assigned the highest fixed weight $w_{\text{ICQ}}$, as they represent the core semantic targets of the \textbf{Edit Instructions}.

\noindent \textbf{Instruction-Dependent Questions (IDQ):} These questions semantically depend on the components or attributes that belong to ICQs. They verify fine-grained spatial or attribute-level consistency, such as shape, texture, or color of certain ICQs, and are only valid if the corresponding ICQ is affirmatively answered. IDQs receive a depth-sensitive dynamic weight:
\begin{equation}
    w_{\text{IDQ}}(l) = 1 + w_{\text{ICQ}} \cdot t_{\text{decay}}^l
\end{equation}
where $l$ is the semantic depth of question j in the dependency graph relative to its corresponding ICQs, and $t_{decay}$ controls the rate of decay across the semantic depth. This mechanism that decays with depth models the intuition that people give priority to the main body of the edit content and tolerate some minor detail errors.

\noindent \textbf{Context-Preserving Questions (CPQ):} These questions that depict non-target editing areas are unrelated to the \textbf{Edit Instructions} but assess whether the semantic content in unedited regions of the image has been preserved. CPQs are assigned a lower constant weight $w_{\text{CPQ}}$, serving as a constraint to verify that unrelated parts of the image remain unchanged. These questions will be used to evaluate whether \textbf{Edit Instructions} are executed correctly and whether irrelevant regions remain unchanged.

\paragraph{Visual Question Answer}
Next, we use a VQA model (we apply Qwen-VL \cite{baiQwenVLVersatileVisionLanguage2023} in this study for its robust zero-shot cross-modal reasoning ability) to answer each question on the edited image. Each response is recorded directly as the categorical label “Yes” or “No,”.

\paragraph{FEditScore Calculation}
Finally, we calculate the final quantitative FEditScore. The overall score is calculated as a normalized weighted sum of accuracies over the three question categories:
\begin{equation}
  \begin{array}{@{}l@{}}
    \text{FEditScore} = \\[1.5ex]
    \displaystyle
    \frac{%
      \sum_{i \in \mathrm{ICQ}} w_{\mathrm{ICQ}}\,\delta_i
    + \sum_{j \in \mathrm{IDQ}} w_{\mathrm{IDQ}}(l_j)\,\delta_j
    + \sum_{k \in \mathrm{CPQ}} w_{\mathrm{CPQ}}\,\delta_k
    }{%
      \sum_{i \in \mathrm{ICQ}} w_{\mathrm{ICQ}}
    + \sum_{j \in \mathrm{IDQ}} w_{\mathrm{IDQ}}(l_j)
    + \sum_{k \in \mathrm{CPQ}} w_{\mathrm{CPQ}}
    }
  \end{array}
\end{equation}
where $i$, $j$, and $k$ indicate individual questions in the ICQ, IDQ, and CPQ set, respectively; $\delta_i = 1$ if the VQA model answers question $i$ as "Yes" and all parent nodes (in the dependency graph) are also "Yes", and $0$ otherwise. We set a threshold $\alpha$ as a filter: any edited image with FEditScore below $\alpha$ is excluded, and those with FEditScore $\alpha$ or above are included in EditGarment.

\begin{figure*}
    \centering
    \includegraphics[width=0.6\linewidth]{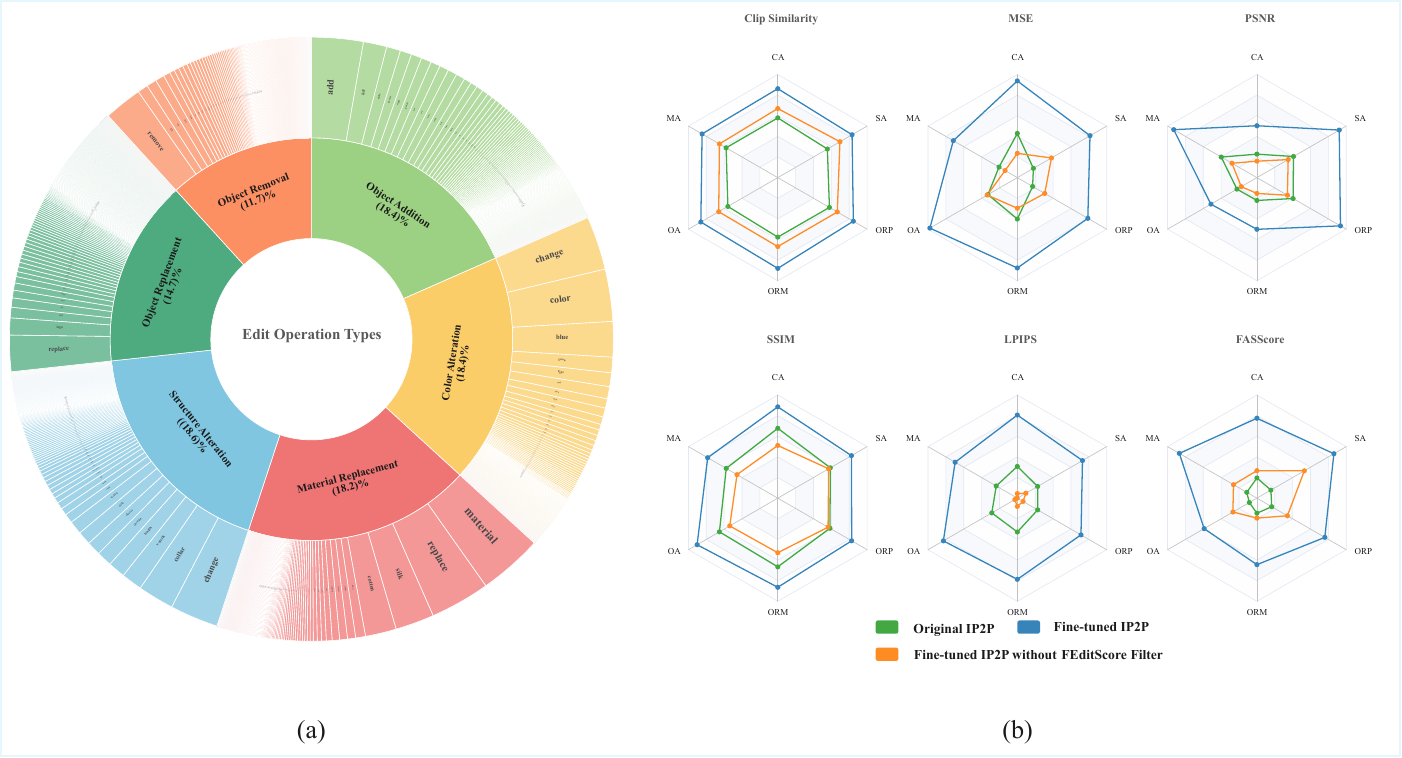}
    \caption{(a) Distribution of edit operation types and keywords. (b) Visualized evaluation comparison between the pre-trained IP2P and IP2P fine-tuned on our dataset (both complete dataset and the version without filter) across six edit types: CA (Color Alteration), MR (Material Replacement), OA (Object Addition), ORP (Object Removal), ORP (Object Replacement), and SA (Structure Alteration). To ensure all metrics follow a consistent ``higher-is-better'' interpretation, we apply inverse normalization to MSE and LPIPS before visualization. }
    \label{fig:3}
\end{figure*}

\subsection{Dataset Analysis}

Base on the 2-stage pipeline, we generate 52,257 edit triplets (original image, edit instructions, edited image) covering six edit types: color alteration, material replacement, structural alteration, object addition, object removal, and object replacement. To construct these triplets, we start from images in the MMDGarment dataset and generate several textual triplets (original description, edit instruction, edited description) per image, ensuring each image contributes to all possible edit types. The distribution of each edit type and high-frequency keywords is shown in Fig. \ref{fig:3} (a). Among these six edit types, structural alteration is the most frequent (18.6\%), while object removal is the least frequent (11.7\%). This indicates a relatively uniform distribution across edit types. After filtering with FEditScore, we retain 20,596 high quality edit triplets with FEditScore less than threshold, namely approximately 39.4 \% of the qualifying data is retained for EditGarment. The minimum resolution in the dataset is 512 × 512. Real garment photos serve as the original images, while edit instructions and edited images are generated by Gemini-Flash \cite{team2023gemini}. A corresponding text description is provided for each sample. 

% EditGarment contains 51,371 triplets of the original image, edit instructions, and the edited image. We define six edit types: color alteration, material replacement, structural alteration, object addition, object removal, and object replacement. To construct these triplets, we started from images in the MMDgarment dataset and generated 1\textasciitilde  2 triplets per image, ensuring each image contributes to all six categories. The distribution of each edit type and high-frequency keywords is shown in Fig. \ref{fig:3} (a). Among these six types, structural alteration is the most frequent (18.6\%), while object removal is the least frequent (11.7\%). This indicates a relatively uniform distribution across edit types.

% The minimum resolution in the dataset is 512 × 512. Real garment photos serve as the original images, while edit instructions and edited images are generated by Gemini-Flash \cite{team2023gemini}. A corresponding text description is provided for each sample. After filtering with FEditScore, approximately 36.7 \% of the qualifying data was retained for EditGarment.

\section{Experiments}
\subsection{Experiment Setup}

\noindent\paragraph{Implementation Details.} 
During dataset construction pipeline, we utilize Qwen-VL-Max-2024-11-19 \cite{baiQwenVLVersatileVisionLanguage2023}, Gemini-2.0-Filash-Preview-Image-Generation \cite{team2023gemini}, and Deepseek R1 \cite{guo2025deepseek}. On average, the generated Original Description consists of 29 words, the Edit Instruction contains 10 words, and the Edited Description comprises 31 words. For FEditScore, we set $w_{CPQ}=1$, $w_{ICQ}=3$, $t_{decay}=0.3$, and filter threshold of FEditScore $\alpha=0.8$ from Section~3.2.

For the instruction-based garment editing task, we adopt InstructPix2Pix (IP2P) \cite{brooks2023instructpix2pix} as our base model, and fine-tune it using our proposed EditGarment dataset. Model training is conducted on 2 $\times$ Nvidia A6000 for over 50 hours with a total of 25,000 training steps and the learning rate 5e-5. During inference, we set the image guidance scale to 1.5, the instruction guidance value to 7.0, and use 20 inference steps. All images are resized to resolution 512 $\times$ 512. Our training set contains 51,357 samples, and our test set consists of 900 samples, evenly distributed across six edit types (150 per type), covering a diverse range of garment modifications. 

\begin{figure}
    \centering
    \includegraphics[width=\linewidth]{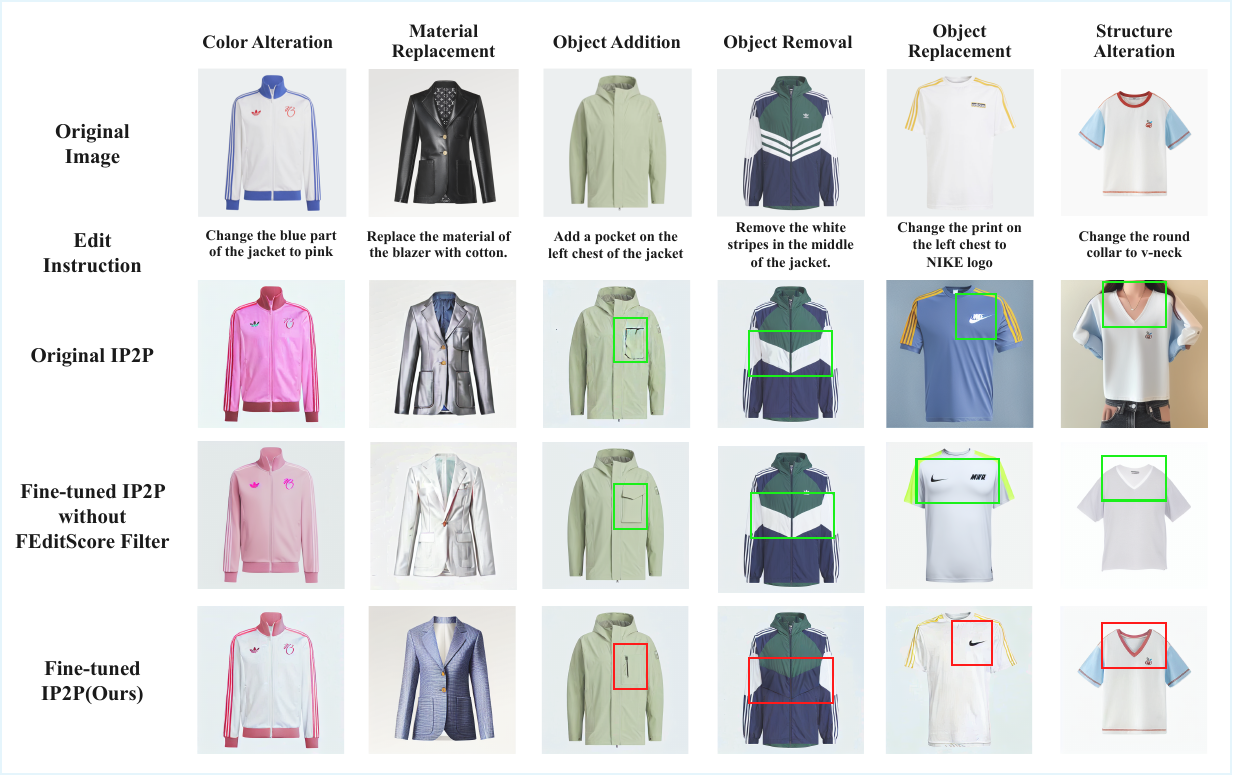}
    \caption{Qualitative comparison between pre-trained IP2P \cite{brooks2023instructpix2pix} and IP2P fine-tuned on our dataset (both complete dataset and the version without evaluation) in six editing categories. \textcolor{green}{Green} boxes indicate sub-optimal editing results, and \textcolor{red}{red} boxes refer to accurate editing results.}
    \label{fig:4}
\end{figure}

\subsection{Qualitative Evaluation}
To provide a baseline for the garment editing task, we train the existing image editing method IP2P \cite{brooks2023instructpix2pix} with EditGarment as shown in Fig. \ref{fig:4}. This comparative analysis shows that EditGarment enables a general image editing model to handle fashion-specific tasks. For example, in the fifth column, the original IP2P model can replace the logo, but cannot guarantee that the unedited regions remain identical to the original image. By contrast, the fine-tuned baseline not only performs general edits such as color modification but also accurately modifies garment attributes such as reshaping collar structures. This result indicates that EditGarment provides the model with rich fashion domain priors. 

\begin{table}[ht]
\centering
\renewcommand{\arraystretch}{1.2}
\setlength{\tabcolsep}{4pt}
\scriptsize
\caption{Quantitative comparison of pre-trained IP2P, IP2P fine-tuned on EditGarment (the complete dataset and the version without FEditScore evaluation) on FEditScore. We choose CLIP similarity \cite{CLIPradfordLearningTransferableVisual2021}, MSE, PSNR, LPIPS \cite{LPIPS-zhang2018unreasonable}, and our introduced FEditScore as evaluation metrics.}
\label{tab:ip2p_comparison}
\begin{tabular}{c|cccccc}
\toprule
\textbf{Method} & \textbf{CLIPSim} $\uparrow$ & \textbf{MSE} $\downarrow$ & \textbf{PSNR} $\uparrow$ & \textbf{SSIM} $\uparrow$ & \textbf{LPIPS} $\downarrow$ & \textbf{FEditScore} $\uparrow$ \\
\midrule
Original IP2P   & 56.372             & 0.279           & 10.017            & 0.596           & 0.382             & 0.082               \\
Ours w/o FEditScore & 65.563         & 0.243           &  7.919            & 0.516           & 0.487             & 0.170               \\
\rowcolor{blue!20}
\textbf{Ours}         & \textbf{84.072}  & \textbf{0.066}  & \textbf{22.991}   & \textbf{0.823}  & \textbf{0.139}    & \textbf{0.425}      \\
\bottomrule
\end{tabular}
\end{table}

\subsection{Quantitative Evaluation}
The quantitative comparison of existing baselines and fine-tuned IP2P is shown in Tab. \ref{tab:ip2p_comparison}. The fine-tuned model performs best in all metrics. Specifically, fine-tuning IP2P with EditGarment yields a 27.7\% increase in CLIP similarity \cite{CLIPradfordLearningTransferableVisual2021} and a 22.7\% increase in SSIM \cite{SSIM} over the IP2P for the overall performance. On FEditScore, our approach achieves a 34.3\% improvement compared to original IP2P. Fig. \ref{fig:3} (b) illustrates visualization comparisons. These results show that EditGarment not only enhances editing quality at the semantic and pixel levels but also enforces correct semantic dependencies throughout the editing process. This comparison confirms that incorporating EditGarment as a baseline improves the performance of general image editing models in the fashion domain.

\subsection{Ablation Study}
We fine-tune IP2P \cite{brooks2023instructpix2pix} with images from our dataset after FEditScore evaluation and images generated directly by MLLM without evaluation, as shown in Tab. \ref{tab:ip2p_comparison} and Fig. \ref{fig:4}. The model trained on the unfiltered image set outperforms the original IP2P on all metrics, but is still behind the version trained on filtered data. This gap shows that our FEditScore evaluation mechanism improves dataset quality by selecting more reliable samples. Unlike traditional metrics such as CLIP similarity or SSIM and VQA-based checks, FEditScore’s semantic-aware evaluation identifies both primary and secondary edit targets. As a result, the model learns to apply the main edits accurately while preserving consistency in non-edited regions. These findings validate the importance of effective supervision for collecting a fine-grained garment image editing dataset.

\subsection{Challenges}
Although experiments demonstrate that EditGarment can serve as a baseline for garment editing, it also introduces new challenges. First, EditGarment contains specialized fashion terms, which may lead to general-purpose image editing models struggling to understand. Existing mechanisms, such as cross-attention, might fail to focus on the unique terms like weave patterns or fabric, causing the model to ignore or misinterpret these instructions. Second, our dataset includes fine-grained detail edits, but diffusion-based methods might lose these details during the editing process because diffusion-based editing pipelines rely on iterative denoising. Fine-grained details such as brand logos or tiny patterns are easily ignored as the model minimizes overall noise. Therefore, the dataset underscores the importance of incorporating domain-specific algorithms and detail-preserving mechanisms in garment editing models.

%\section{Limitations}
%数据集的limitation。现实应用可能涉及到同时出现多个编辑指令。

\section{Conclusion}
In this paper, we propose EditGarment, the first and largest multimodal dataset for garment image editing to the best of our knowledge. EditGarment contains 20,596 edits, each consisting of a real-world garment image, an edit instruction, and the edited image. To build this dataset, we propose an automated pipeline with six predefined edit types corresponding to fashion design requirements. This approach both fills gaps in real-world garment distributions and mitigates biases in MLLM-generated data. To evaluate data quality, we introduce Fashion Edit Score (FEditScore), a dependency graph based metric that detects hallucinations and enforces semantic consistency. Experiments show that models fine-tuned on EditGarment achieve impressive performance on garment editing tasks, demonstrating the value of our dataset and evaluation framework. EditGarment also introduces new challenges, such as handling specialized fashion instructions and preserving fine-grained details, which remain for future research to discover.

\begin{acks}
This work is supported by the Key Technology R\&D Program of Ningbo (2023Z143, 2025Z028), the Natural Science Foundation of the Jiangsu Higher Education Institutions of China (23KJB120013), and Xi’an Jiaotong-Liverpool University Teaching Development Fund (TDF23/24-R27-222)
\end{acks}

\bibliographystyle{ACM-Reference-Format}
\bibliography{main}


\begin{thebibliography}{40}

%%% the Web address.

\ifx \showCODEN    \undefined \def \showCODEN     #1{\unskip}     \fi
\ifx \showISBNx    \undefined \def \showISBNx     #1{\unskip}     \fi
\ifx \showISBNxiii \undefined \def \showISBNxiii  #1{\unskip}     \fi
\ifx \showISSN     \undefined \def \showISSN      #1{\unskip}     \fi
\ifx \showLCCN     \undefined \def \showLCCN      #1{\unskip}     \fi
\ifx \shownote     \undefined \def \shownote      #1{#1}          \fi
\ifx \showarticletitle \undefined \def \showarticletitle #1{#1}   \fi
\ifx \showURL      \undefined \def \showURL       {\relax}        \fi
% invisible to TeX
\providecommand\bibfield[2]{#2}
\providecommand\bibinfo[2]{#2}
\providecommand\natexlab[1]{#1}
\providecommand\showeprint[2][]{arXiv:#2}

\bibitem[Bai et~al\mbox{.}(2023)]%
        {baiQwenVLVersatileVisionLanguage2023}
\bibfield{author}{\bibinfo{person}{Jinze Bai}, \bibinfo{person}{Shuai Bai}, \bibinfo{person}{Shusheng Yang}, \bibinfo{person}{Shijie Wang}, \bibinfo{person}{Sinan Tan}, \bibinfo{person}{Peng Wang}, \bibinfo{person}{Junyang Lin}, \bibinfo{person}{Chang Zhou}, {and} \bibinfo{person}{Jingren Zhou}.} \bibinfo{year}{2023}\natexlab{}.
\newblock \showarticletitle{Qwen-{{VL}}: {{A Versatile Vision-Language Model}} for {{Understanding}}, {{Localization}}, {{Text Reading}}, and {{Beyond}}}.
\newblock \bibinfo{journal}{\emph{arXiv preprint arXiv:2312.11805}}.
\newblock


\bibitem[Baldrati et~al\mbox{.}(2023)]%
        {baldratiMultimodalGarmentDesigner2023}
\bibfield{author}{\bibinfo{person}{Alberto Baldrati}, \bibinfo{person}{Davide Morelli}, \bibinfo{person}{Giuseppe Cartella}, \bibinfo{person}{Marcella Cornia}, \bibinfo{person}{Marco Bertini}, {and} \bibinfo{person}{Rita Cucchiara}.} \bibinfo{year}{2023}\natexlab{}.
\newblock \showarticletitle{Multimodal {{Garment Designer}}: {{Human-Centric Latent Diffusion Models}} for {{Fashion Image Editing}}}.
\newblock \bibinfo{journal}{\emph{In Proceedings of CVPR}}, \bibinfo{pages}{23393--23402}.
\newblock


\bibitem[Baldrati et~al\mbox{.}(2024)]%
        {baldratiMultimodalConditionedLatentDiffusion2024}
\bibfield{author}{\bibinfo{person}{Alberto Baldrati}, \bibinfo{person}{Davide Morelli}, \bibinfo{person}{Marcella Cornia}, \bibinfo{person}{Marco Bertini}, {and} \bibinfo{person}{Rita Cucchiara}.} \bibinfo{year}{2024}\natexlab{}.
\newblock \showarticletitle{Multimodal-conditioned latent diffusion models for fashion image editing}.
\newblock \bibinfo{journal}{\emph{arXiv preprint arXiv:2403.14828}}.
\newblock


\bibitem[Bodur et~al\mbox{.}(2024)]%
        {bodurIEditLocalisedTextguided2024}
\bibfield{author}{\bibinfo{person}{Rumeysa Bodur}, \bibinfo{person}{Erhan Gundogdu}, \bibinfo{person}{Binod Bhattarai}, \bibinfo{person}{Tae-Kyun Kim}, \bibinfo{person}{Michael Donoser}, {and} \bibinfo{person}{Loris Bazzani}.} \bibinfo{year}{2024}\natexlab{}.
\newblock \showarticletitle{{{iEdit}}: {{Localised Text-guided Image Editing}} with {{Weak Supervision}}}.
\newblock \bibinfo{journal}{\emph{In Proceedings of CVPR Workshop}}, \bibinfo{pages}{7426--7435}.
\newblock


\bibitem[Brack et~al\mbox{.}(2024)]%
        {brackLEDITSLimitlessImage2024}
\bibfield{author}{\bibinfo{person}{Manuel Brack}, \bibinfo{person}{Felix Friedrich}, \bibinfo{person}{Katharina Kornmeier}, \bibinfo{person}{Linoy Tsaban}, \bibinfo{person}{Patrick Schramowski}, \bibinfo{person}{Kristian Kersting}, {and} \bibinfo{person}{Apolin{\'a}rio Passos}.} \bibinfo{year}{2024}\natexlab{}.
\newblock \showarticletitle{{{LEDITS}}++: {{Limitless Image Editing Using Text-to-Image Models}}}.
\newblock \bibinfo{journal}{\emph{In Proceedings of CVPR}}, \bibinfo{pages}{8861--8870}.
\newblock


\bibitem[Brooks et~al\mbox{.}(2023)]%
        {brooks2023instructpix2pix}
\bibfield{author}{\bibinfo{person}{Tim Brooks}, \bibinfo{person}{Aleksander Holynski}, {and} \bibinfo{person}{Alexei~A Efros}.} \bibinfo{year}{2023}\natexlab{}.
\newblock \showarticletitle{Instructpix2pix: Learning to follow image editing instructions}.
\newblock \bibinfo{journal}{\emph{Proceedings of the CVPR}}, \bibinfo{pages}{18392--18402}.
\newblock


\bibitem[Brown et~al\mbox{.}(2020)]%
        {brown2020language}
\bibfield{author}{\bibinfo{person}{Tom Brown}, \bibinfo{person}{Benjamin Mann}, \bibinfo{person}{Nick Ryder}, \bibinfo{person}{Melanie Subbiah}, \bibinfo{person}{Jared~D Kaplan}, \bibinfo{person}{Prafulla Dhariwal}, \bibinfo{person}{Arvind Neelakantan}, \bibinfo{person}{Pranav Shyam}, \bibinfo{person}{Girish Sastry}, \bibinfo{person}{Amanda Askell}, {et~al\mbox{.}}} \bibinfo{year}{2020}\natexlab{}.
\newblock \showarticletitle{Language models are few-shot learners}.
\newblock \bibinfo{journal}{\emph{In Proceedings of NeurIPS}}  \bibinfo{volume}{33}, \bibinfo{pages}{1877--1901}.
\newblock


\bibitem[Chakrabarty et~al\mbox{.}(2023)]%
        {chakrabartyLearningFollowObjectCentric2023a}
\bibfield{author}{\bibinfo{person}{Tuhin Chakrabarty}, \bibinfo{person}{Kanishk Singh}, \bibinfo{person}{Arkadiy Saakyan}, {and} \bibinfo{person}{Smaranda Muresan}.} \bibinfo{year}{2023}\natexlab{}.
\newblock \showarticletitle{Learning to {{Follow Object-Centric Image Editing Instructions Faithfully}}}.
\newblock \bibinfo{journal}{\emph{In Proceedings of EMNLP}}.
\newblock


\bibitem[Cho et~al\mbox{.}(2023)]%
        {cho2023davidsonian}
\bibfield{author}{\bibinfo{person}{Jaemin Cho}, \bibinfo{person}{Yushi Hu}, \bibinfo{person}{Roopal Garg}, \bibinfo{person}{Peter Anderson}, \bibinfo{person}{Ranjay Krishna}, \bibinfo{person}{Jason Baldridge}, \bibinfo{person}{Mohit Bansal}, \bibinfo{person}{Jordi Pont-Tuset}, {and} \bibinfo{person}{Su Wang}.} \bibinfo{year}{2023}\natexlab{}.
\newblock \showarticletitle{Davidsonian scene graph: Improving reliability in fine-grained evaluation for text-to-image generation}.
\newblock \bibinfo{journal}{\emph{In Proceedings of ICLR}}.
\newblock


\bibitem[Choi et~al\mbox{.}(2021)]%
        {VITON-HD_Choi_2021_CVPR}
\bibfield{author}{\bibinfo{person}{Seunghwan Choi}, \bibinfo{person}{Sunghyun Park}, \bibinfo{person}{Minsoo Lee}, {and} \bibinfo{person}{Jaegul Choo}.} \bibinfo{year}{2021}\natexlab{}.
\newblock \showarticletitle{VITON-HD: High-Resolution Virtual Try-On via Misalignment-Aware Normalization}.
\newblock \bibinfo{journal}{\emph{In Proceedings of CVPR}}, \bibinfo{pages}{14131--14140}.
\newblock


\bibitem[Dhariwal and Nichol(2021)]%
        {dhariwalDiffusionModelsBeat}
\bibfield{author}{\bibinfo{person}{Prafulla Dhariwal} {and} \bibinfo{person}{Alex Nichol}.} \bibinfo{year}{2021}\natexlab{}.
\newblock \showarticletitle{Diffusion {{Models Beat GANs}} on {{Image Synthesis}}}.
\newblock \bibinfo{journal}{\emph{In Proceedings of NeurIPS}}  \bibinfo{volume}{34}, \bibinfo{pages}{8780--8794}.
\newblock


\bibitem[Guo et~al\mbox{.}(2025a)]%
        {guo2025deepseek}
\bibfield{author}{\bibinfo{person}{Daya Guo}, \bibinfo{person}{Dejian Yang}, \bibinfo{person}{Haowei Zhang}, \bibinfo{person}{Junxiao Song}, \bibinfo{person}{Ruoyu Zhang}, \bibinfo{person}{Runxin Xu}, \bibinfo{person}{Qihao Zhu}, \bibinfo{person}{Shirong Ma}, \bibinfo{person}{Peiyi Wang}, \bibinfo{person}{Xiao Bi}, {et~al\mbox{.}}} \bibinfo{year}{2025}\natexlab{a}.
\newblock \showarticletitle{Deepseek-r1: Incentivizing reasoning capability in llms via reinforcement learning}.
\newblock \bibinfo{journal}{\emph{arXiv preprint arXiv:2501.12948}}.
\newblock


\bibitem[Guo et~al\mbox{.}(2025b)]%
        {guo2025higarmentcrossmodalharmonybased}
\bibfield{author}{\bibinfo{person}{Junyi Guo}, \bibinfo{person}{Jingxuan Zhang}, \bibinfo{person}{Fangyu Wu}, \bibinfo{person}{Huanda Lu}, \bibinfo{person}{Qiufeng Wang}, \bibinfo{person}{Wenmian Yang}, \bibinfo{person}{Eng~Gee Lim}, {and} \bibinfo{person}{Dongming Lu}.} \bibinfo{year}{2025}\natexlab{b}.
\newblock \showarticletitle{HiGarment: Cross-modal Harmony Based Diffusion Model for Flat Sketch to Realistic Garment Image}.
\newblock \bibinfo{journal}{\emph{arXiv preprint arXiv:2505.23186}}.
\newblock


\bibitem[Hu et~al\mbox{.}(2023)]%
        {huTIFAAccurateInterpretable2023}
\bibfield{author}{\bibinfo{person}{Yushi Hu}, \bibinfo{person}{Benlin Liu}, \bibinfo{person}{Jungo Kasai}, \bibinfo{person}{Yizhong Wang}, \bibinfo{person}{Mari Ostendorf}, \bibinfo{person}{Ranjay Krishna}, {and} \bibinfo{person}{Noah~A. Smith}.} \bibinfo{year}{2023}\natexlab{}.
\newblock \showarticletitle{{{TIFA}}: {{Accurate}} and {{Interpretable Text-to-Image Faithfulness Evaluation}} with {{Question Answering}}}.
\newblock \bibinfo{journal}{\emph{In Proceedings of ICCV}}, \bibinfo{pages}{20406--20417}.
\newblock


\bibitem[Huang et~al\mbox{.}(2025)]%
        {huang2025diffusion}
\bibfield{author}{\bibinfo{person}{Yi Huang}, \bibinfo{person}{Jiancheng Huang}, \bibinfo{person}{Yifan Liu}, \bibinfo{person}{Mingfu Yan}, \bibinfo{person}{Jiaxi Lv}, \bibinfo{person}{Jianzhuang Liu}, \bibinfo{person}{Wei Xiong}, \bibinfo{person}{He Zhang}, \bibinfo{person}{Liangliang Cao}, {and} \bibinfo{person}{Shifeng Chen}.} \bibinfo{year}{2025}\natexlab{}.
\newblock \showarticletitle{Diffusion model-based image editing: A survey}.
\newblock \bibinfo{journal}{\emph{IEEE Transactions on Pattern Analysis and Machine Intelligence}}.
\newblock


\bibitem[Huang et~al\mbox{.}(2024)]%
        {huangSmartEditExploringComplex2024}
\bibfield{author}{\bibinfo{person}{Yuzhou Huang}, \bibinfo{person}{Liangbin Xie}, \bibinfo{person}{Xintao Wang}, \bibinfo{person}{Ziyang Yuan}, \bibinfo{person}{Xiaodong Cun}, \bibinfo{person}{Yixiao Ge}, \bibinfo{person}{Jiantao Zhou}, \bibinfo{person}{Chao Dong}, \bibinfo{person}{Rui Huang}, \bibinfo{person}{Ruimao Zhang}, {and} \bibinfo{person}{Ying Shan}.} \bibinfo{year}{2024}\natexlab{}.
\newblock \showarticletitle{{{SmartEdit}}: {{Exploring Complex Instruction-Based Image Editing}} with {{Multimodal Large Language Models}}}.
\newblock \bibinfo{journal}{\emph{In Proceedings of CVPR}}, \bibinfo{pages}{8362--8371}.
\newblock


\bibitem[{Huberman-Spiegelglas} et~al\mbox{.}(2024)]%
        {huberman-spiegelglasEditFriendlyDDPM2024}
\bibfield{author}{\bibinfo{person}{Inbar {Huberman-Spiegelglas}}, \bibinfo{person}{Vladimir Kulikov}, {and} \bibinfo{person}{Tomer Michaeli}.} \bibinfo{year}{2024}\natexlab{}.
\newblock \showarticletitle{An {{Edit Friendly DDPM Noise Space}}: {{Inversion}} and {{Manipulations}}}.
\newblock \bibinfo{journal}{\emph{In Proceedings of CVPR}}, \bibinfo{pages}{12469--12478}.
\newblock


\bibitem[Hui et~al\mbox{.}(2024)]%
        {huiHQEditHighQualityDataset2024}
\bibfield{author}{\bibinfo{person}{Mude Hui}, \bibinfo{person}{Siwei Yang}, \bibinfo{person}{Bingchen Zhao}, \bibinfo{person}{Yichun Shi}, \bibinfo{person}{Heng Wang}, \bibinfo{person}{Peng Wang}, \bibinfo{person}{Yuyin Zhou}, {and} \bibinfo{person}{Cihang Xie}.} \bibinfo{year}{2024}\natexlab{}.
\newblock \showarticletitle{HQ-Edit: {A} High-Quality Dataset for Instruction-based Image Editing}.
\newblock \bibinfo{journal}{\emph{In Proceedings of ICLR}}.
\newblock


\bibitem[Liu et~al\mbox{.}(2024)]%
        {liuUnderstandingCrossSelfAttention2024}
\bibfield{author}{\bibinfo{person}{Bingyan Liu}, \bibinfo{person}{Chengyu Wang}, \bibinfo{person}{Tingfeng Cao}, \bibinfo{person}{Kui Jia}, {and} \bibinfo{person}{Jun Huang}.} \bibinfo{year}{2024}\natexlab{}.
\newblock \showarticletitle{Towards {{Understanding Cross}} and {{Self-Attention}} in {{Stable Diffusion}} for {{Text-Guided Image Editing}}}.
\newblock \bibinfo{journal}{\emph{In Proceedings of CVPR}}, \bibinfo{pages}{7817--7826}.
\newblock


\bibitem[Mishkin et~al\mbox{.}(2022)]%
        {dalle-mishkin2022risks}
\bibfield{author}{\bibinfo{person}{Pamela Mishkin}, \bibinfo{person}{Lama Ahmad}, \bibinfo{person}{Miles Brundage}, \bibinfo{person}{Gretchen Krueger}, {and} \bibinfo{person}{Girish Sastry}.} \bibinfo{year}{2022}\natexlab{}.
\newblock \bibinfo{booktitle}{\emph{DALL·E 2 Preview - Risks and Limitations}}.
\newblock
\urldef\tempurl%
\url{https://github.com/openai/dalle-2-preview/blob/main/system-card.md}
\showURL{%
\tempurl}


\bibitem[Mokady et~al\mbox{.}(2023)]%
        {mokadyNulltextInversionEditing2023}
\bibfield{author}{\bibinfo{person}{Ron Mokady}, \bibinfo{person}{Amir Hertz}, \bibinfo{person}{Kfir Aberman}, \bibinfo{person}{Yael Pritch}, {and} \bibinfo{person}{Daniel {Cohen-Or}}.} \bibinfo{year}{2023}\natexlab{}.
\newblock \showarticletitle{Null-Text {{Inversion}} for {{Editing Real Images}} Using {{Guided Diffusion Models}}}.
\newblock \bibinfo{journal}{\emph{In Proceedings of CVPR}}, \bibinfo{pages}{6038--6047}.
\newblock


\bibitem[Morelli et~al\mbox{.}(2023)]%
        {morelliLaDIVTONLatentDiffusion2023}
\bibfield{author}{\bibinfo{person}{Davide Morelli}, \bibinfo{person}{Alberto Baldrati}, \bibinfo{person}{Giuseppe Cartella}, \bibinfo{person}{Marcella Cornia}, \bibinfo{person}{Marco Bertini}, {and} \bibinfo{person}{Rita Cucchiara}.} \bibinfo{year}{2023}\natexlab{}.
\newblock \showarticletitle{{{LaDI-VTON}}: {{Latent Diffusion Textual-Inversion Enhanced Virtual Try-On}}}.
\newblock \bibinfo{journal}{\emph{In Proceedings of ACM MM}}, \bibinfo{pages}{8580--8589}.
\newblock


\bibitem[Morelli et~al\mbox{.}(2022)]%
        {morelliDressCodeHighResolution2022}
\bibfield{author}{\bibinfo{person}{Davide Morelli}, \bibinfo{person}{Matteo Fincato}, \bibinfo{person}{Marcella Cornia}, \bibinfo{person}{Federico Landi}, \bibinfo{person}{Fabio Cesari}, {and} \bibinfo{person}{Rita Cucchiara}.} \bibinfo{year}{2022}\natexlab{}.
\newblock \showarticletitle{Dress {{Code}}: {{High-Resolution Multi-Category Virtual Try-On}}}.
\newblock \bibinfo{journal}{\emph{In Proceedings of CVPR}}.
\newblock


\bibitem[Mu et~al\mbox{.}(2025)]%
        {muEditARUnifiedConditional2025}
\bibfield{author}{\bibinfo{person}{Jiteng Mu}, \bibinfo{person}{Nuno Vasconcelos}, {and} \bibinfo{person}{Xiaolong Wang}.} \bibinfo{year}{2025}\natexlab{}.
\newblock \showarticletitle{{{EditAR}}: {{Unified Conditional Generation}} with {{Autoregressive Models}}}.
\newblock \bibinfo{journal}{\emph{arXiv preprint arXiv:2501.04699}}.
\newblock


\bibitem[OpenAI(2024)]%
        {openaiGPT4TechnicalReport2024}
\bibfield{author}{\bibinfo{person}{OpenAI}.} \bibinfo{year}{2024}\natexlab{}.
\newblock \showarticletitle{{{GPT-4 Technical Report}}}.
\newblock \bibinfo{journal}{\emph{arXiv preprint arXiv:2303.08774}}.
\newblock


\bibitem[Peebles and Xie(2023)]%
        {peeblesScalableDiffusionModels2023}
\bibfield{author}{\bibinfo{person}{William Peebles} {and} \bibinfo{person}{Saining Xie}.} \bibinfo{year}{2023}\natexlab{}.
\newblock \showarticletitle{Scalable {{Diffusion Models}} with {{Transformers}}}.
\newblock \bibinfo{journal}{\emph{arXiv preprint arXiv:2212.09748}}, \bibinfo{pages}{4195--4205}.
\newblock


\bibitem[Radford et~al\mbox{.}(7 24)]%
        {CLIPradfordLearningTransferableVisual2021}
\bibfield{author}{\bibinfo{person}{Alec Radford}, \bibinfo{person}{Jong~Wook Kim}, \bibinfo{person}{Chris Hallacy}, \bibinfo{person}{Aditya Ramesh}, \bibinfo{person}{Gabriel Goh}, \bibinfo{person}{Sandhini Agarwal}, \bibinfo{person}{Girish Sastry}, \bibinfo{person}{Amanda Askell}, \bibinfo{person}{Pamela Mishkin}, \bibinfo{person}{Jack Clark}, \bibinfo{person}{Gretchen Krueger}, {and} \bibinfo{person}{Ilya Sutskever}.} \bibinfo{year}{2021-07-18/2021-07-24}\natexlab{}.
\newblock \showarticletitle{Learning {{Transferable Visual Models From Natural Language Supervision}}}.
\newblock \bibinfo{journal}{\emph{In Proceedings of ICML}}, \bibinfo{pages}{8748--8763}.
\newblock


\bibitem[Rombach et~al\mbox{.}(2022)]%
        {rombach2022high}
\bibfield{author}{\bibinfo{person}{Robin Rombach}, \bibinfo{person}{Andreas Blattmann}, \bibinfo{person}{Dominik Lorenz}, \bibinfo{person}{Patrick Esser}, {and} \bibinfo{person}{Bj{\"o}rn Ommer}.} \bibinfo{year}{2022}\natexlab{}.
\newblock \showarticletitle{High-resolution image synthesis with latent diffusion models}.
\newblock \bibinfo{journal}{\emph{In Proceedings of CVPR}}, \bibinfo{pages}{10684--10695}.
\newblock


\bibitem[Sheynin et~al\mbox{.}(2024)]%
        {sheyninEmuEditPrecise2024}
\bibfield{author}{\bibinfo{person}{Shelly Sheynin}, \bibinfo{person}{Adam Polyak}, \bibinfo{person}{Uriel Singer}, \bibinfo{person}{Yuval Kirstain}, \bibinfo{person}{Amit Zohar}, \bibinfo{person}{Oron Ashual}, \bibinfo{person}{Devi Parikh}, {and} \bibinfo{person}{Yaniv Taigman}.} \bibinfo{year}{2024}\natexlab{}.
\newblock \showarticletitle{Emu {{Edit}}: {{Precise Image Editing}} via {{Recognition}} and {{Generation Tasks}}}.
\newblock \bibinfo{journal}{\emph{In Proceedings of CVPR}}, \bibinfo{pages}{8871--8879}.
\newblock


\bibitem[Shuai et~al\mbox{.}(2024)]%
        {shuaiSurveyMultimodalGuidedImage2024}
\bibfield{author}{\bibinfo{person}{Xincheng Shuai}, \bibinfo{person}{Henghui Ding}, \bibinfo{person}{Xingjun Ma}, \bibinfo{person}{Rongcheng Tu}, \bibinfo{person}{Yu-Gang Jiang}, {and} \bibinfo{person}{Dacheng Tao}.} \bibinfo{year}{2024}\natexlab{}.
\newblock \showarticletitle{A survey of multimodal-guided image editing with text-to-image diffusion models}.
\newblock \bibinfo{journal}{\emph{arXiv preprint arXiv:2406.14555}}.
\newblock


\bibitem[Team et~al\mbox{.}(2023)]%
        {team2023gemini}
\bibfield{author}{\bibinfo{person}{Gemini Team}, \bibinfo{person}{Rohan Anil}, \bibinfo{person}{Sebastian Borgeaud}, \bibinfo{person}{Jean-Baptiste Alayrac}, \bibinfo{person}{Jiahui Yu}, \bibinfo{person}{Radu Soricut}, \bibinfo{person}{Johan Schalkwyk}, \bibinfo{person}{Andrew~M Dai}, \bibinfo{person}{Anja Hauth}, \bibinfo{person}{Katie Millican}, {et~al\mbox{.}}} \bibinfo{year}{2023}\natexlab{}.
\newblock \showarticletitle{Gemini: a family of highly capable multimodal models}.
\newblock \bibinfo{journal}{\emph{arXiv preprint arXiv:2312.11805}}.
\newblock


\bibitem[Tewel et~al\mbox{.}(2024)]%
        {tewelAdditTrainingFreeObject2024}
\bibfield{author}{\bibinfo{person}{Yoad Tewel}, \bibinfo{person}{Rinon Gal}, \bibinfo{person}{Dvir Samuel}, \bibinfo{person}{Yuval Atzmon}, \bibinfo{person}{Lior Wolf}, {and} \bibinfo{person}{Gal Chechik}.} \bibinfo{year}{2024}\natexlab{}.
\newblock \showarticletitle{Add-It: {{Training-Free Object Insertion}} in {{Images With Pretrained Diffusion Models}}}.
\newblock \bibinfo{journal}{\emph{arXiv preprint arXiv:2411.07232}}.
\newblock


\bibitem[Wang and Ye(2024)]%
        {wangTexFitTextDrivenFashion2024}
\bibfield{author}{\bibinfo{person}{Tongxin Wang} {and} \bibinfo{person}{Mang Ye}.} \bibinfo{year}{2024}\natexlab{}.
\newblock \showarticletitle{{{TexFit}}: {{Text-Driven Fashion Image Editing}} with {{Diffusion Models}}}.
\newblock \bibinfo{journal}{\emph{In Proceedings of AAAI}} \bibinfo{volume}{38}, \bibinfo{number}{9}, \bibinfo{pages}{10198--10206}.
\newblock


\bibitem[Wang et~al\mbox{.}(2004)]%
        {SSIM}
\bibfield{author}{\bibinfo{person}{Zhou Wang}, \bibinfo{person}{A.C. Bovik}, \bibinfo{person}{H.R. Sheikh}, {and} \bibinfo{person}{E.P. Simoncelli}.} \bibinfo{year}{2004}\natexlab{}.
\newblock \showarticletitle{Image quality assessment: from error visibility to structural similarity}.
\newblock \bibinfo{journal}{\emph{IEEE Transactions on Image Processing}} \bibinfo{volume}{13}, \bibinfo{number}{4}, \bibinfo{pages}{600--612}.
\newblock


\bibitem[Zhang et~al\mbox{.}(2023a)]%
        {zhangMAGICBRUSHManuallyAnnotated}
\bibfield{author}{\bibinfo{person}{Kai Zhang}, \bibinfo{person}{Lingbo Mo}, \bibinfo{person}{Wenhu Chen}, \bibinfo{person}{Huan Sun}, {and} \bibinfo{person}{Yu Su}.} \bibinfo{year}{2023}\natexlab{a}.
\newblock \showarticletitle{{{MAGICBRUSH}} : {{A Manually Annotated Dataset}} for {{Instruction-Guided Image Editing}}}. \bibinfo{pages}{31428--31449}.
\newblock


\bibitem[Zhang et~al\mbox{.}(2018)]%
        {LPIPS-zhang2018unreasonable}
\bibfield{author}{\bibinfo{person}{Richard Zhang}, \bibinfo{person}{Phillip Isola}, \bibinfo{person}{Alexei~A Efros}, \bibinfo{person}{Eli Shechtman}, {and} \bibinfo{person}{Oliver Wang}.} \bibinfo{year}{2018}\natexlab{}.
\newblock \showarticletitle{The unreasonable effectiveness of deep features as a perceptual metric}.
\newblock \bibinfo{journal}{\emph{In Proceedings of CVPR}}, \bibinfo{pages}{586--595}.
\newblock


\bibitem[Zhang et~al\mbox{.}(2024a)]%
        {zhangGarmentAlignerTexttoGarmentGeneration2024}
\bibfield{author}{\bibinfo{person}{Shiyue Zhang}, \bibinfo{person}{Zheng Chong}, \bibinfo{person}{Xujie Zhang}, \bibinfo{person}{Hanhui Li}, \bibinfo{person}{Yuhao Cheng}, \bibinfo{person}{Yiqiang Yan}, {and} \bibinfo{person}{Xiaodan Liang}.} \bibinfo{year}{2024}\natexlab{a}.
\newblock \showarticletitle{{{GarmentAligner}}: {{Text-to-Garment Generation}} via {{Retrieval-augmented Multi-level Corrections}}}.
\newblock \bibinfo{journal}{\emph{In Proceedings of ECCV}}, \bibinfo{pages}{148--164}.
\newblock


\bibitem[Zhang et~al\mbox{.}(2024b)]%
        {zhangHIVEHarnessingHuman2024}
\bibfield{author}{\bibinfo{person}{Shu Zhang}, \bibinfo{person}{Xinyi Yang}, \bibinfo{person}{Yihao Feng}, \bibinfo{person}{Can Qin}, \bibinfo{person}{Chia-Chih Chen}, \bibinfo{person}{Ning Yu}, \bibinfo{person}{Zeyuan Chen}, \bibinfo{person}{Huan Wang}, \bibinfo{person}{Silvio Savarese}, \bibinfo{person}{Stefano Ermon}, \bibinfo{person}{Caiming Xiong}, {and} \bibinfo{person}{Ran Xu}.} \bibinfo{year}{2024}\natexlab{b}.
\newblock \showarticletitle{{{HIVE}}: {{Harnessing Human Feedback}} for {{Instructional Visual Editing}}}.
\newblock \bibinfo{journal}{\emph{In Proceedings of CVPR}}, \bibinfo{pages}{9026--9036}.
\newblock


\bibitem[Zhang et~al\mbox{.}(2022)]%
        {zhangARMANIPartlevelGarmentText2022}
\bibfield{author}{\bibinfo{person}{Xujie Zhang}, \bibinfo{person}{Yu Sha}, \bibinfo{person}{Michael~C. Kampffmeyer}, \bibinfo{person}{Zhenyu Xie}, \bibinfo{person}{Zequn Jie}, \bibinfo{person}{Chengwen Huang}, \bibinfo{person}{Jianqing Peng}, {and} \bibinfo{person}{Xiaodan Liang}.} \bibinfo{year}{2022}\natexlab{}.
\newblock \showarticletitle{{{ARMANI}}: {{Part-level Garment-Text Alignment}} for {{Unified Cross-Modal Fashion Design}}}.
\newblock \bibinfo{journal}{\emph{In Proceedings of ACM MM}}, \bibinfo{pages}{4525--4535}.
\newblock


\bibitem[Zhang et~al\mbox{.}(2023b)]%
        {zhangDiffClothDiffusionBased2023}
\bibfield{author}{\bibinfo{person}{Xujie Zhang}, \bibinfo{person}{Binbin Yang}, \bibinfo{person}{Michael~C. Kampffmeyer}, \bibinfo{person}{Wenqing Zhang}, \bibinfo{person}{Shiyue Zhang}, \bibinfo{person}{Guansong Lu}, \bibinfo{person}{Liang Lin}, \bibinfo{person}{Hang Xu}, {and} \bibinfo{person}{Xiaodan Liang}.} \bibinfo{year}{2023}\natexlab{b}.
\newblock \showarticletitle{{{DiffCloth}}: {{Diffusion Based Garment Synthesis}} and {{Manipulation}} via {{Structural Cross-modal Semantic Alignment}}}.
\newblock \bibinfo{journal}{\emph{In Proceedings of ICCV}}, \bibinfo{pages}{23154--23163}.
\newblock


\end{thebibliography}

%\appendix






\end{document}